\pdfoutput=1

\documentclass[11pt]{article}

\usepackage{acl}

\usepackage{placeins}

\usepackage{times}
\usepackage{latexsym}

\usepackage[T1]{fontenc}

\usepackage[utf8]{inputenc}

\usepackage{microtype}

\usepackage{multirow}
\usepackage{float}
\usepackage{multicol}
\usepackage{amsmath}
\usepackage{cleveref}
\usepackage{graphicx} 

\usepackage{array}
\usepackage{ragged2e}
\newcolumntype{M}[1]{>{\RaggedRight\hspace{0pt}}m{#1}}
\newcolumntype{P}[1]{>{\RaggedRight\hspace{0pt}}p{#1}}

\newcommand\blfootnote[1]{%
  \begingroup
  \renewcommand\thefootnote{}\footnote{#1}%
  \addtocounter{footnote}{-1}%
  \endgroup
}

\title{Understanding Metrics for Paraphrasing}

\author{Omkar Patil, Rahul Singh \and Tarun Joshi \\
  Corporate Model Risk, Wells Fargo \\
  \texttt{\{omkar.d.patil, rahul.singh3, tarun.joshi1\}@wellsfargo.com} \\}

\begin{document}
\maketitle
\begin{abstract}
Paraphrase generation is a difficult problem. This is not only because of the limitations in text generation capabilities but also due that to the lack of a proper definition of what qualifies as a paraphrase and corresponding metrics to measure how good it is. Metrics for evaluation of paraphrasing quality is an on going research problem. Most of the existing metrics in use having been borrowed from other tasks do not capture the complete essence of a good paraphrase, and often fail at borderline-cases. In this work, we propose a novel metric $ROUGE_P$ to measure the quality of paraphrases along the dimensions of adequacy, novelty and fluency. We also provide empirical evidence to show that the current natural language generation metrics are insufficient to measure these desired properties of a good paraphrase. We look at paraphrase model fine-tuning and generation from the lens of metrics to gain a deeper understanding of what it takes to generate and evaluate a good paraphrase.
\end{abstract}

\section{Introduction}
\blfootnote{The views expressed in the paper are those of the authors and do not represent the views of Wells Fargo.}
The task of sentential paraphrasing is to generate paraphrases for a given sentence. 
In loose terms, two sentences can be called paraphrases of each other if they convey the same meaning in different words. 
The generated paraphrase should preserve the semantics of the original sentence. Moreover, it should also avoid appearing similar to the original sentence in terms of choice of words and/or sentence structure.
Paraphrasing has found several applications, notably in fields like data augmentation \cite{hegde2020unsupervised} and robustness testing \cite{iyyer2018adversarial}.
Several paraphrasing models have been proposed based on LSTM \cite{hochreiter1997long}, autoencoders \cite{liou2014autoencoder} and transformers\cite{vaswani2017attention}. 
Several metrics too have been adapted or proposed, but there is no consensus on criteria for evaluation of paraphrase quality. In this work, we propose that generated paraphrases be evaluated based on their adequacy, novelty, fluency and correctness. The interpretation of adequacy adapted for paraphrasing is to measure the degree of semantics preserved in the generated paraphrase relative to the gold-standard reference or source \cite{DBLP:journals/corr/abs-2006-14799}. Fluency can be understood as a measure to quantify how devoid the generation is of repetition, spelling and grammatical mistakes \cite{DBLP:journals/corr/abs-2006-14799}. Novelty can be gauged by how different the generated paraphrase is from the input to the model. Finally, correctness can be used to check whether the generated paraphrase has any information that contradicts the input sentence or hallucinations which are out of scope of the input \cite{https://doi.org/10.48550/arxiv.2005.00661}.

\noindent In this paper, we present our findings on how good paraphrases can be generated and evaluated. We analyze the fine-tuning and generation process using metrics and reveal various trade-offs in the process. Shortcomings of existing metrics in common generation settings leads us to propose a new metric $ROUGE_P$. These generation settings are used to create challenging paraphrase examples to showcase the weakness of current metrics used for paraphrase evaluation. 
Our main contributions are- 

 \begin{itemize} 
 \item Propose a novel metric for evaluating the quality of paraphrases based on ROUGE \cite{lin2004rouge}, incorporating adequacy, novelty and fluency aspects into it.


 \item Quantify vocabulary diversity in model output and understand how model fine-tuning affects diversity, and adequacy and novelty of generated paraphrases.

 \item Utilize a novel selection metric for candidate paraphrases to leverage the trade-off between their adequacy and novelty.

 \item Demonstrate the shortcomings of popular metrics in current literature such as BLEU\cite{papineni2002bleu}, PINC \cite{chen2011collecting} and TER \cite{snover2006study} when applied for paraphrase evaluation.
 \end{itemize}
 
The paper is structured as follows. In \Cref{lit:review}, we present a brief literature review on the common metrics that have been used for evaluating paraphrases, followed by a review of papers that have used GPT-2 for the task of paraphrase generation. \Cref{rouge:p}, presents details on our novel metric $ROUGE_P$. We present our methodology of paraphrase model development in \Cref{meth} and analysis of paraphrase evaluation using $ROUGE_P$ in \Cref{rouge:action}. We discuss how metrics helps us better analyze model fine-tuning and generation in \Cref{anal:metrics}. In \Cref{rev:para}, we demonstrate generation settings where existing paraphrase evaluation metrics fail. We discuss the limitations of our work in \Cref{limitations}. Finally, \Cref{conclusions} provides a summarization of our contributions and directions for future work.

\section{Literature Review}
\label{lit:review}

\subsection{Metrics}
\label{metrics:intro}
Several metrics have been used for automatic evaluation of model generated paraphrases with reference during testing. 
Metrics like BLEU \cite{papineni2002bleu}, METEOR \cite{lavie2007meteor}, ROUGE \cite{lin2004rouge} and TER \cite{snover2006study}
are intended to measure the adequacy of generated paraphrases. PINC \cite{chen2011collecting} and selfBLEU \cite{zhu2018texygen} were designed to gauge the novelty and diversity of generated paraphrases respectively. PEM \cite{liu2010pem} was created to incorporate adequacy, novelty and fluency aspects for phrasal paraphrasing. \Cref{app:metrics:desc} provides details on these and other common metrics used in paraphrasing literature. In the following work, \textit{ref} refers to the reference sentence and \textit{cand} refers to one of the many candidate paraphrases generated from the model. \textit{gen} refers to the final paraphrase sentence selected from the candidate set as the model output.
The same metric may be calculated in different ways depending upon the setting.
Often, metrics are to be interpreted at the corpus level and not for 
individual sentence (\textit{snt}) pairs, which is likewise marked with \textit{corpus}. 
Most datasets have pairs of sentences which are paraphrases of each other, referred to as S1 
and S2, where S1 is fed into the model and the generated output \textit{gen} may be evaluated with either S1 or S2. 
But, some datasets like MSCOCO captions \cite{chen2015microsoft} offer multiple sentences as paraphrases of each other.
For MSCOCO, the input to the model would be S1 and \textit{gen} would be compared with S1 or S2, S3, S4, S5.

\subsection{Models}
Several models, including some based on GPT-2 \cite{radford2019language} have been proposed for the task of generating paraphrases. 
GPT-2 has good language generation capabilities. \citet{witteveen2019paraphrasing} use 
GPT-2 to generate candidate paraphrases which are then filtered using sentence similarity score \cite{cer2018universal} and $ROUGE_L$ with the input sentence. 
\citet{hegde2020unsupervised} fine-tune GPT-2 in an unsupervised manner with the input to the model as a corrupted form of the output. 
The input is corrupted by removing the stop words, random shuffling and synonym substitution. Paraphrases are generated through the model and 
filtered using sentence similarity with the input.

\section{ROUGE-P: A novel metric for paraphrase evaluation}
\label{rouge:p}

Addressing how semantically similar two sentences are is a subset of the problem of how good a paraphrase is. 
Other than being semantically similar, a paraphrase also needs to be distinct from the input sentence.
This factor of novelty is especially important while reporting adequacy scores, 
as simple greedy decoding results in very high scores due to parroting, as seen in \Cref{msr:top}.
Most previous papers \cite{gupta2018deep,prakash2016neural,hegde2020unsupervised} do not report how novel their paraphrases are while reporting adequacy scores.
This idea is in line with \citet{chen2011collecting} and \citet{Xu-EtAl-2014:TACL} who use
BLEU as a metric for adequacy and PINC for lexical dissimilarity. However, \citet{shen2022revisiting}
show negative correlation for BLEU with human judgement on adequacy for paraphrasing, and we show that PINC
fails as a dissimilarity metric for \textit{reversed paraphrases} in \Cref{rev:para}. \citet{liu2010pem} propose a single metric 
to account for both adequacy and novelty at phrasal level paraphrasing. To that end, we propose $ROUGE_P$ (\Cref{rp}), a ROUGE based
paraphrasing metric to account for adequacy ($srcROUGE_1$), novelty (\Cref{nf}) and fluency (\Cref{ff}). We note that for the rest of the paper, all metrics with \textit{src} affixed in front are to be calculated with the input S1 (srcPINC is written as PINC itself). 

\small
\begin{align}
&nf = \left(1- \left(\frac{max(srcROUGE_L-benchROUGE_L, 0)}{1-benchROUGE_L}\right)^\beta\right) \label{nf} \\
&ff = \left(1- \left(\frac{max(benchROUGE_L-srcROUGE_L, 0)}{benchROUGE_L}\right)^\gamma\right) \label{ff} \\
&lenpen = min(1, exp(1-\frac{gen\ length}{src\ length})) \label{lenpen} \\
&ROUGE_P = srcROUGE_1*nf*ff*lenpen \label{rp}
\end{align}
\normalsize

\begin{figure}
\centering
  \centering
  \includegraphics[width=1\linewidth]{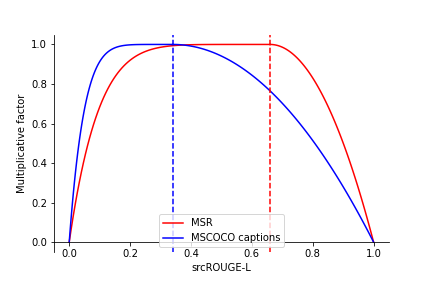}
  \captionof{figure}{The effect of novelty and fluency factor is shown for a range of values of $srcROUGE_L$. 
  The red and blue dotted lines indicate $benchROUGE_L$ for MSR and MSCOCO captions respectively.
  The novelty and fluency factors are inactive before and after the dotted lines respectively. 
  $\beta$ is taken as 2 to limit the penalization for the initial 10\% of the remaining range of $srcROUGE_L$ 
  after the dotted line. $\gamma$ is taken as 7 to start effectively penalizing generation for non-fluency only when $srcROUGE_L$ drops below the 50\%
  mark of $benchROUGE_L$. 
  }
  \label{metric:pen}
\end{figure}

\noindent Here, $benchROUGE_L$ or benchmark $ROUGE_L$ corresponds to microaverage of $ROUGE_L$ of the
reference paraphrases with the source, for the whole document/corpus.
$nf$ or novelty factor in the equation accounts for novelty when $srcROUGE_L$ exceeds the benchmark $ROUGE_L$.
It does that by penalizing $srcROUGE_1$ at a polynomial rate, bringing it down to 0
in the extreme case of pure parroting.  $ff$ or fluency factor is to prevent the metric from scoring high when $ROUGE_L$
drops very low but $ROUGE_1$ is still high, as could happen for unfluent or jumbled sentences. $ff$ brings down the score
at a much lower rate initially than $nf$, for $srcROUGE_L$ lower than the benchmark.
$lenpen$ or the length penalty is to prevent generation longer than the input to score high due to the recall based adequacy measure.
$gen$ and $src$ in $lenpen$ correspond to the generated and source sentences.
$\beta$ is a constant taken as 2, based on a heuristic to limit the penalization to 0.99 for $\frac{1-benchROUGE_L}{10}$ of excess 
$srcROUGE_L$ over the benchmark. $\gamma$ is taken as 7 to limit the penalization to 0.99 for when $srcROUGE_L$ drops to 
$\frac{benchROUGE_L}{2}$. There is very little penalty on $ROUGE_1$ when $srcROUGE_L$ 
hovers around the document benchmark, and has positive correlation with human judgement on adequacy \cite{shen2022revisiting}.
\Cref{metric:pen} shows how novelty and fluency factor change with respect to the dataset benchmarks.
The corpus $ROUGE_P$ can be calculated as average of individual sentence values of the metric.

\section{Methodology and Experimentation}
\label{meth}
\subsection{Fine-tuning and Generating Paraphrases using GPT-2}
Output text in GPT-2 is produced one token at a time, auto-regressively; that is, at each time step a token will be produced using previous tokens as context. 
Since we are dealing with directed text generation \cite{holtzman2019curious}, we also need to keep the input sentence S1 in the context of each token generation step. 
To fine-tune the model on a paraphrasing dataset, we structure the input sequence as [EOS]S1[SEP]S2[EOS], for each paraphrase pair (S1, S2). 
At each time-step, GPT-2 generates a probability distribution $P$ over its internal vocabulary based on the token 
at the previous time step and all the preceding ones. The loss at each time step is calculated as 
the cross-entropy between the probability distribution produced and the one-hot vector of the next token in the sequence. 
More details on the model-fine-tuning process can be found in \Cref{app:a}.

We fine-tune GPT-2 small (110M parameters) and medium (345M parameters) on MSR \cite{dolan2004unsupervised} and MSCOCO captions \cite{chen2015microsoft} training datasets. All the models are fine-tuned for 10 epochs under AdamW\cite{loshchilov2018fixing} 
with a constant learning rate of either 1e-4 or 1e-5 and a weight decay of 0.01. We use a Tesla V100-SXM2-32GB GPU for this purpose.

At inference, the input to the model is [EOS]S1[SEP]. We keep generating tokens till a pre-decided length 
or truncate the output after [EOS] token is generated. At each step the model produces a probability 
distribution over its vocabulary and a token is sampled from it as the output. 
Beam search is often used as a proxy for generating most likely sentences. The alternative is to disregard the sentence probabilities and sample individual tokens from the distribution at each time-step. Top-k, top-p \cite{holtzman2019curious} and temperature scaling help us to sample tokens effectively from the distributions. More details on the generation process can be found in \Cref{app:b}.

\subsection{Selection and Evaluation metrics}
Either through beam search or by repeating the whole process of sequential decoding multiple times,
several candidate paraphrases can be generated for an input. 
We refer to selection metrics as those which are used to select the final model output from the
generated candidates. 
We utilize a weighted harmonic mean of $srcROUGE_1$ and $1-srcROUGE_L$ scores of the candidate as a selection metric:
\begin{equation}
\frac{srcROUGE_1*(1-srcROUGE_L)*w}{srcROUGE_1+(1-srcROUGE_L)*w}
\label{sel:metric}
\end{equation}
Here the weight $w$ controls how much importance we give to $srcROUGE_1$,
with higher values prioritizing adequacy over novelty, with an upper and lower limit on $srcROUGE_L$. We also add a length penalty similar to BLEU.

Evaluation metrics are used to evaluate the final model generated paraphrase with the reference or source.
Paraphrases should be evaluated on semantic adequacy, novelty, fluency and correctness. 
Along with our proposed metric $ROUGE_P$, we implement several other popular metrics in literature for the purpose of comparison and analysis.
\citet{shen2022revisiting} show high correlation for $srcROUGE_L$ with human judgement in paraphrasing.
For adequacy we also calculate BLEU (and TER) score between the generated paraphrase and the references.
We calculate BLEU (and TER) using the implementation provided by \citet{post-2018-call}. 
The calculation details of most metrics for multiple references at a corpus level (as compared to sentence level) are provided in \Cref{pop:metrics}.
We refer to parroting as when the model output is exactly similar or only slightly diffferent from its input S1\cite{mao2019polly}. 
For certain datasets, parroting is an essential consideration while generating and scoring paraphrases, as parroted 
output will score high, despite not being meaningful. Since very high values of $srcROUGE_L$ correspond to low novelty of generated paraphrases, 
we benchmark the metric to the value obtained for paraphrases within the dataset ($ROUGE_L$ between source and reference).
For novelty we show the average of f-measure of $srcROUGE_L$ (and srcBLEU, and average of 
PINC scores) for the testing corpus. Although most generations are fluent, perplexity can be used as a proxy for fluency of paraphrases.
Automatic evaluation of correctness of the paraphrase (distinct from semantic adequacy) 
is left for future work.

\subsection{Datasets}
We use MSR paraphrase dataset  \cite{dolan2004unsupervised} and MSCOCO captions \cite{chen2015microsoft} for our experiments. 
We train and test the model separately on both of them.
MSR is a small dataset with very close paraphrases created by scraping off news sources from the web. We follow the default test-train 
split for our purpose, resulting in 2.7K and 1.1K pairs for training and evaluation respectively. On the other hand MSCOCO
captions is sizeable, diverse and with multiple references (5 captions for each image) for each input.
For fine-tuning, one caption among five is randomly deleted and the model is trained on two pairs formed out of the remaining four. 
During evaluation, to use multiple references, we pick one caption as the input 
and use the other four as references for comparison with the model generated paraphrase. 
We have 331K training pairs and 40K set of five sentences for evaluation in this dataset.
For selected results, we randomly sample and use 5\% of the test data set in MSCOCO captions,
indicated as MSCOCO*.

\section{$ROUGE_P$ in Action}
\label{rouge:action}
\Cref{mscoco:top} and \Cref{msr:top}
show the result of various sampling methods. `std. in srcRL' stands for standard deviation in $srcROUGE_L$. 
The results are for learning rate of 1e-4 and top-k with $k=5$ and top-p with $p=0.95$. 
In \Cref{mscoco:top}, the first row refers to the metrics calculated between the paraphrases present in the dataset, with $S1$ denoting the input that goes into the model, 
and $S2,\ S3,\ S4$ and $S5$ being the reference ones. For the first row, BLEU is calculated as multi-reference metrics between S1 and the reference paraphrases.
$srcROUGE_L$, std. in srcRL, $srcROUGE_1$ and $ROUGE_P$ are the average of the metrics between each of the reference paraphrases and $S1$. 
In \Cref{msr:top}, random sampling helps reduce $srcROUGE_L$, but 
the standard deviation in $srcROUGE_L$ values is very high for MSR, implying that few sentences are very similar to the input,
and few are very further from it, which is also undesirable. As can be seen, $ROUGE_P$ penalizes sentences with higher $srcROUGE_L$ 
than the benchmark, resulting in decoding configurations having lower score despite a high $srcROUGE_1$ or BLEU.
For results on MSCOCO captions in \Cref{mscoco:top}, top-k and top-p sampling bring the $srcROUGE_L$ levels closer to the dataset benchmark, 
but result in lower adequacy scores on metrics like BLEU and $srcROUGE_1$. $ROUGE_P$ still indicates a higher score for 
greedy in \Cref{mscoco:top} as the higher $srcROUGE_L$ than benchmark does not offset the higher unigram overlap.
\Cref{para:samp} shows some paraphrases for MSCOCO and MSR generated using top-p and greedy sampling.

\begin{table*}[!ptbh]
\caption{Effect of greedy, top-k and top-p sampling methods on MSCOCO for GPT-2 small and medium.
These results need to be interpreted in the light of loose nature of paraphrases within the MSCOCO captions dataset.
Hence, greedy decoding performs favourably with higher $ROUGE_P$ scores than even the benchmark.
The higher $srcROUGE_L$ scores relative to the benchmark do not offset higher unigram overlap of the generated paraphrases with the source.}
\vspace{-5mm}
\begin{center}    
    \begin{tabular}{ | m{2cm} | c | l | l | l | l | l | l |}
    \hline
    \multicolumn{2}{|c|}{Model} & BLEU & $srcROUGE_1$ &  $srcROUGE_L$ & Std. in srcRL & $ROUGE_P$ \\ \hline
    \multicolumn{2}{|c|}{S1 and (S2, S3, S4, S5)} & \textbf{19.55} & \textbf{0.39} & \textbf{0.34} & \textbf{0.16} & \textbf{0.33} \\ \hline

    \multirow{2}{*}{GPT-2 small} & Greedy & 26.71 & 0.51 & 0.47 & 0.19 & 0.39 \\ \cline{2-7}
						 & Top-k=5 & 16.93 & 0.45 & 0.39 & 0.17 & 0.36 \\ \cline{2-7}
						& Top-p=0.95 & 13.64 & 0.41 & 0.36 & 0.16 & 0.34 \\ \hline

   \multirow{2}{*}{GPT-2 med} & Greedy & 22.36 & 0.47 & 0.43 & 0.17 & 0.38 \\ \cline{2-7}
						& Top-k=5 & 15.7 & 0.42 & 0.37 & 0.16 & 0.35 \\ \cline{2-7}
						& Top-p=0.95 & 14 & 0.40 & 0.35 & 0.16 & 0.34 \\ \hline
    \end{tabular}
\end{center}
\label{mscoco:top}
\end{table*}

\begin{table*}[!ptbh]
\caption{Effect of greedy, random, top-k and top-p sampling methods on MSR for GPT-2 small and medium.
We see higher values of $srcROUGE_L$ in greedy for both GPT-2 small and medium indicating partial parroting.
This issue is ameliorated by random, top-k and top-p sampling, but they are still plagued by high standard deviation in 
$srcROUGE_L$ values, indicating inconsistent generation. Our metric $ROUGE_P$ is also sensitive to this as the metric is calculated at a sentence level relative to a corpus-wide benchmark.}
\vspace{-5mm}
\begin{center}    
    \begin{tabular}{ | m{2cm} | c | l | l | l | l | l | l |}
    \hline
    \multicolumn{2}{|c|}{Model} & BLEU & $srcROUGE_1$ &  $srcROUGE_L$ & Std. in srcRL & $ROUGE_P$ \\ \hline
    \multicolumn{2}{|c|}{S1 and S2 in MSR} & \textbf{47.45} & \textbf{0.71} & \textbf{0.66} & \textbf{0.13} & \textbf{0.60} \\ \hline

    \multirow{2}{*}{GPT-2 small} & Greedy & 39.33 & 0.79 & 0.77 & 0.22 & 0.42 \\ \cline{2-7}
						 & Sampling & 33.64 & 0.70 & 0.67 & 0.23 & 0.46 \\ \cline{2-7}
						 & Top-k=5 & 34.22 & 0.72 & 0.7 & 0.22 & 0.46 \\ \cline{2-7}
						 & Top-p=0.95 & 34.53 & 0.72 & 0.69 & 0.23 & 0.45 \\ \hline

   \multirow{2}{*}{GPT-2 med}   & Greedy & 39.36 & 0.78 & 0.77 & 0.21 & 0.40 \\ \cline{2-7}
   						& Sampling & 36.52 & 0.74 & 0.72 & 0.22 & 0.45 \\ \cline{2-7}
						& Top-k=5 & 36.75 & 0.74 & 0.72 & 0.22 & 0.43 \\ \cline{2-7}
						& Top-p=0.95 & 37.11 & 0.75 & 0.73 & 0.22 & 0.43 \\ \hline
    \end{tabular}
\end{center}
\label{msr:top}
\end{table*}

\begin{figure*}
\centering
\begin{minipage}{.5\textwidth}
  \centering
  \includegraphics[width=1\linewidth]{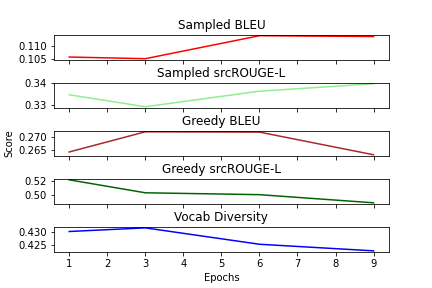}
\end{minipage}%
\begin{minipage}{.5\textwidth}
  \centering
  \includegraphics[width=1\linewidth]{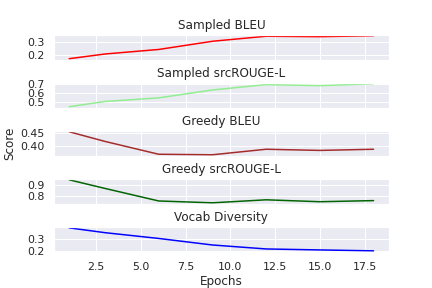}
\end{minipage}
  \captionof{figure}{Variation in test metrics with the number of epochs 
  GPT-2 was fine-tuned for on MSCOCO (left) and MSR (right). Greedy sampling results of the model are a good indicator of 
  what it has learnt during its fine-tuning. Greedy $srcROUGE_L$ drops 
  (from much higher than the benchmark) as the training progresses, implying that the model is shedding its tendency to parrot.
  Vocabulary diversity also dips as the model is fine-tuned further to generated sharper distributions. 
  Note that metrics for sampled generation improve with further fine-tuning.}
  \label{anal:fin}
\end{figure*}

\section{Understanding the Trade-offs Using Metrics}
\label{anal:metrics}
\subsection{Analyzing Model Fine-tuning}
We would like to quantify certain characteristics of the model output
to understand how fine-tuning affects paraphrase generation capabilities. We introduce vocabulary diversity, to quantify the 
capacity of a model to produce lexically diverse paraphrases for a single input under specified decoding configuration.
Vocabulary diversity \cite{richards1987type} is calculated as number of unique tokens together in source, reference and 10 paraphrases 
sampled from the model divided by the total number of tokens in them. 
The results in \Cref{anal:fin} were obtained for GPT-2 fine-tuned with a learning rate of 1e-4 and evaluated on MSCOCO* captions and MSR.
Although we overfit the models, the resulting distributions that they learn yield better performance when sampled from.
As shown by the $srcROUGE_L$ curves for greedy selection, it also helps the model escape parroting.
The fine-tuning trade-off with adequacy is that the model loses it's capacity to generate diverse paraphrases for an input sentence as the distribution sharpens, as shown by the downward diversity curves in \Cref{anal:fin}. 
Such a metrics-based analysis can help us set a stopping criteria for fine-tuning, balancing the required adequacy and diversity in the generated paraphrases. Note that diversity can be increased with temperature scaling of the generated distributions, which may sometimes yield higher adequacy scores for the same diversity.

\subsection{Influencing Paraphrase Generation}
Metrics can be used as filters to choose paraphrases with desired properties from the many candidates generated by the model.
Results from such a candidate selection process are shown in \Cref{msr:cand} and \Cref{mscoco:cand}. Ten paraphrase candidates are generated for 
each sentence using sampling. We provide results for two values of $w$ in \Cref{sel:metric}, 1.5 and 3. Sampling for MSR in \Cref{msr:cand}
results in paraphrases with high variance in quality, which is also reflected in the low $ROUGE_P$ score. 
Candidate selection helps to not only maintain the consistency 
of paraphrases that are generated but also allows us to leverage the adequacy-novelty trade-off. 
Overall, candidate selection yields higher $ROUGE_P$ scores than simple sampling or greedy decoding.
\Cref{para:samp} shows some paraphrases for MSR generated using top-p followed by candidate selection.

\begin{table*}
\caption{Candidate selection results for MSR test dataset using GPT-2 fine-tuned at a learning rate of 1e-4 for 10 epochs.
Not only does filtering using our metric \Cref{sel:metric} achieve higher $ROUGE_P$ scores, it does so consistently across sentences, as shown by the standard deviation in $srcROUGE_L$. Another interesting use can be to generate paraphrases of desired novelty, trading-off adequacy for it using the parameter $w$ in the selection metric. }

\begin{center}    
    \begin{tabular}{ | m{2cm} | m{1.4cm} | l | l | l | l | l |}
    \hline
    Model & Decoding & BLEU & $srcROUGE_1$ &  $srcROUGE_L$ & \textit{Std. in srcRL} & $ROUGE_P$ \\ \hline
    \multicolumn{2}{|c|}{S1 and S2 in MSR} & \textbf{47.45} & \textbf{0.71} & \textbf{0.66} & \textbf{0.13} & \textbf{0.60} \\ \hline

    \multirow{3}{*}{GPT-2 small} & Sampling & 33.64 & 0.7 & 0.67 & 0.23 & 0.46\\ \cline{2-7}
						 & w=1.5     & 28.2 & 0.66 & 0.61 & 0.12 & 0.59 \\ \cline{2-7}
						 & w=3        & 30.08 & 0.69 & 0.64 & 0.11 & 0.62 \\ \cline{1-7}

    \multirow{3}{*}{GPT-2 med} & Sampling & 36.52 & 0.74 & 0.72 & 0.22 & 0.45\\ \cline{2-7}
						 & w=1.5     & 32.7 & 0.69 & 0.65 & 0.14 & 0.57 \\ \cline{2-7}
						 & w=3        & 33.79 & 0.72 & 0.68 & 0.13 & 0.59 \\ \cline{1-7}

    \end{tabular}
\end{center}
\label{msr:cand}
\end{table*}

\begin{table*}
\caption{Candidate selection results for MSCOCO* using GPT-2 fine-tuned at a learning rate of 1e-4 for 5 epochs.
Again, results for MSCOCO need to be interpreted in the light of the loose nature of paraphrases within the MSCOCO captions dataset.
The $srcROUGE_L$ after filtering is higher than our baseline, but not very high in absolute terms. $ROUGE_P$ does end up doing much better than the baseline and random sampling due to more than commensurate increase in unigram overlaps.}

\begin{center}    
    \begin{tabular}{ | m{2cm} | m{1.4cm} | l | l | l | l | l |}
    \hline
    Model & Decoding & BLEU & $srcROUGE_1$ & $srcROUGE_L$ & Std. in srcRL & $ROUGE_P$ \\ \hline
    \multicolumn{2}{|c|}{S1 and (S2, S3, S4, S5)} & \textbf{19.43} & \textbf{0.39} & \textbf{0.34} & \textbf{0.15} & \textbf{0.33} \\ \hline

    \multirow{3}{*}{GPT-2 small} & Sampling & 11.4 & 0.37 & 0.33 & 0.17 & 0.32 \\ \cline{2-7}
						 & w=1.5     & 14.78 & 0.5 & 0.46 & 0.16 & 0.43 \\ \cline{2-7}
						 & w=3        & 15.46 & 0.51 & 0.47 & 0.17 & 0.44 \\ \cline{1-7}

    \multirow{3}{*}{GPT-2 med} & Sampling & 10.88 & 0.38 & 0.33 & 0.17 & 0.32 \\ \cline{2-7}
						 & w=1.5     & 14.96 & 0.5 & 0.45 & 0.16 & 0.43 \\ \cline{2-7}
						 & w=3        & 15.6 & 0.51 & 0.47 & 0.17 & 0.43 \\ \cline{1-7}

    \end{tabular}
\end{center}
\label{mscoco:cand}
\end{table*}

\begin{table*}
\small
\caption{Normal and reversed paraphrases generated using beam search for MSR. Our proposed metric $ROUGE_P$ performs as per expectation on challenger paraphrases. Many of the conventional metrics for adequacy and novelty do not respond well to reversed generation. Last row shows results for GPT-2 small trained on MSR at a learning rate of 1e-5 for 10 epochs. This generation is intended to mimic the input sentences very closely. $ROUGE_P$ assigns a low score to quality of paraphrases due to the lack of novelty in them.}
\begin{center}    
    \begin{tabular}{ | m{2.3cm} | m{1.2cm} | m{1.2cm} | m{1.2cm} | m{1.2cm} | m{1.2cm} | m{1.2cm} | m{1.2cm} | m{1.2cm} | m{1.2cm} |}
    \hline
    \multicolumn{2}{|c|}{Model} & BLEU & TER & src ROUGE\textsubscript{1} & src ROUGE\textsubscript{L} & Std. in srcRL & PINC & ROUGE\textsubscript{P} \\ \hline
    \multicolumn{2}{|c|}{S1 and S2 in MSR} & \textbf{47.45} & \textbf{49.63} & \textbf{0.71} & \textbf{0.66} & \textbf{0.13} & \textbf{0.52} & \textbf{0.60} \\ \hline

    \multirow{2}{*}{GPT-2 small} & Normal & 31.7 & 66.28 & 0.71 & 0.688 & 0.11 & 0.424  & 0.63\\ \cline{2-9}
						& Reversed & 30.45 & 69.92 & 0.76 & 0.629 & 0.19 & 0.422 & 0.62 \\ \hline

   \multirow{2}{*}{GPT-2 med} & Normal & 32.28 & 64.63 & 0.70 & 0.685 & 0.12 & 0.417  & 0.64\\ \cline{2-9}
						& Reversed &32.76 & 67.73 &  0.78 & 0.681 & 0.18 & 0.392 & 0.60\\ \hline
	
   \multicolumn{2}{|c|}{GPT-2 small with greedy sampling} & 43.66 & 53.38 & 0.90 & 0.90 & 0.15 & 0.14 & 0.23 \\ \hline
    \end{tabular}
\end{center}
\label{msr:beam} \normalsize
\end{table*}

\section{Creation of Challenging Paraphrases for Evaluation} \label{rev:para}
Paraphrasing model development reveals many scenarios where metric scores do not do justice to the generated paraphrases.
We showcase various natural generation settings where the conventional metrics might fail.
To create challenging paraphrase examples, we generate reversed paraphrases for each of the input sentences from the MSR test dataset. We do so using beam search, the results for which are shown in \Cref{msr:beam}. 
As beam search also involves generation of multiple candidates and their consequent filtering, the results follow a similar pattern 
as candidate selection after sequential decoding.
`Normal' here refers to paraphrases which are not biased towards a specific goal. 
We have used a beam size of 20 for these experiments. Based on our observations, it also becomes imperative to apply a moving window repetition
penalty for beam search, which in this case is a penalty of 5 for a window of 40 previous tokens. 
For generating reversed paraphrases, we influence the probability distribution at each step to increase the probability of tokens at the other end of the sentence.
This does not conform to a specific linguistic type and it is not necessary that all the samples within the dataset be reversed.
Parroting or novelty is no longer a concern now, but rather preservation of the meaning of S1 by candidate paraphrases.
Hence, for selecting from multiple candidates, 
we choose the candidate with highest cosine similarity between its BERT\cite{devlin2018bert} embedding of that of the input paraphrase.

We can observe several anomalies for the metric scores of reversed paraphrases. 
Where a higher PINC score indicates more dissimilarity in \Cref{msr:beam}, the PINC score stays the same and decreases for reversed paraphrases 
generated by GPT-2 small and medium respectively, in comparison to `normal'. As expected, the effect of reversing is visible in the $srcROUGE_L$ scores,
where reversed paraphrases have a lower $srcROUGE_L$ score with S1. This shows that PINC may not accurately quantify the novelty for very diverse paraphrases. 
TER has been used by many previous works \cite{gupta2018deep,prakash2016neural,hegde2020unsupervised} to measure the adequacy of their generated paraphrases.
Where a lower TER with S2 indicates a better paraphrase match with the reference, results for reversed paraphrases have 
higher than expected TER, despite having comparable BLEU to`normal' paraphrases. This is in line with the intuition that edit-based metrics may indicate higher number of edits being
required for a very novel paraphrase, which may in fact be better. Thus, TER and other edit-based metrics may fail for very novel paraphrase pairs as measures of adequacy, especially when just a single reference is present.
$ROUGE_P$ scores are only slightly higher for reversed paraphrases due to the higher variance in novelty despite having a lower mean of $srcROUGE_L$ than normal paraphrases.
\Cref{para:samp} shows some examples of reversed paraphrases.

As is evident from \Cref{anal:fin}, the paraphrase model tends to parrot the input sentence if not fine-tuned properly. We use this to develop challenging examples where the generated paraphrases are very similar to the input sentences. This is evident from the $srcROUGE_L$ scores for the last row in \Cref{msr:beam}. BLEU and TER indicate high quality paraphrases, which is clearly not the case. $ROUGE_P$ correctly assigns a very
low score, arising from the novelty factor penalizing the unigram overlap. \Cref{para:samp} has samples for a generation setting which leads to near parroting on MSR. This goes on to show that the adequacy of paraphrases must be interpreted in the light of their novelty, which our metric $ROUGE_P$ is designed to do.

\section{Limitations}
\label{limitations}
The proposed metric $ROUGE_P$ uses two constants $\beta$ and $\gamma$ which are set according to heuristics. For future work, they could be calibrated to human judgement which also takes novelty of the paraphrases into account. Further work can also be done to find better alternatives for $ROUGE_1$, especialy taking semantics into consideration. Better alternatives also exist for estimation of fluency of a sentence. In this paper, we have used ROUGE-based measures to keep the simplicity of the metric intact.

\section{Conclusions and Future Work}
\label{conclusions}
In this work, we propose a novel metric $ROUGE_P$ which takes the adequacy, novelty and fluency of the generated paraphrase into account, while being simple to use. To the best of our knowledge, there has been no such metric proposed for a composite judgement on the quality of our model sentential paraphrases.
The inspiration for the metric arose from failure of existing metrics to properly evaluate the quality of generated paraphrases. We present these challenging paraphrase examples and showcase how existing metrics fail on them.
Model paraphrase generation was analyzed from a metrics point of view to reveal a trade-off between adequacy and novelty, which forms the backbone of our proposed metric.

For future, we would like to work towards testing the robustness of text-classification models using paraphrased inputs 
and generating specific kinds of paraphrases using language models.
Another emerging direction is automatic evaluation of the correctness of paraphrases, a critical aspect not covered by adequacy.

\clearpage


\clearpage
\appendix

\section{Metrics for Paraphrasing}
\label{app:metrics:desc}
In \Cref{pop:metrics}, \textit{ref} refers to the reference sentence and \textit{cand} refers to one of the many candidate paraphrases generated from the model. \textit{gen} would then refer to the final paraphrase sentence selected from the candidate set as the model output.
Often, metrics are to be interpreted at the corpus level and not for 
individual sentence (\textit{snt}) pairs, which is likewise marked with \textit{corpus}. 
Evaluation of metrics for multiple references is also given in the description wherever applicable.

\begin{table*}
\caption{Description of some common metrics used in paraphrase generation literature. 
We also specify the evaluation details at a corpus level wherever applicable. See \Cref{metrics:intro}
for an introduction.}
\begin{center}
    \begin{tabular}{ | p{15cm} |}    
    \hline
    Metrics\\ \hline
    \textbf{BLEU}\cite{papineni2002bleu} - Bilingual Evaluation Understudy\\
 $BLEU_{corpus} = min(1, exp(1-\frac{ref\ length}{gen\ length})) * (\prod_{i=1}^{4}{precision_i})^{0.25}$ \\
 $precision_i =  \frac{\sum_{snt \in gen-corpus}{\sum_{i-gram \in snt}{count_{clip}(i-gram)}}}{\sum_{snt \in gen-corpus}{\sum_{i-gram \in snt}{count_{gen}(i-gram)}}}$ \\
 where $count_{clip} = min(count_{gen}, count_{ref})$\\
 The first part of $BLEU_{corpus}$ is the brevity penalty and the second part is the geometric mean of modified n-gram precision scores.  
 In case of multiple references, the closest in length is used to calculate the brevity penalty 
 and the count is clipped at the maximum count of the i-gram in a single reference. \\ \hline
						
    \textbf{METEOR}\cite{lavie2007meteor} - Metric for Evaluation for Translation with Explicit Ordering\\ 
 Unigram mapping (alignment) between two strings is created using exact, porter stem and synonymy. Based on the word mapping,
 a parametrized harmonic mean of unigram precision and recall is calculated. \\
 $F_{mean} = \frac{PR}{\alpha P + (1-\alpha)R}$ \\
Where $P = \frac{Mapped\ unigrams}{Unigrams\ in\ gen}$ and  $R = \frac{Mapped\ unigrams}{Unigrams\ in\ ref}$ \\
$METEOR = (1-Pen)*F_{mean}$ where $Pen$ is the alignment penalty.\\
For multiple references: $METEOR_{multi} = max_i(METEOR(reference_i, candidate))$ \\ \hline

    \textbf{ROUGE}\cite{lin2004rouge} - Recall Oriented Understudy for Gisting Evaluation\\ 
 $ROUGE_N = \frac{\sum_{snt \in ref-corpus}{\sum_{n-gram \in snt}{count_{match}(n-gram)}}}{\sum_{snt \in ref-corpus}{\sum_{n-gram \in snt}{count_{ref}(n-gram)}}}$ \\
 $ROUGE_N$ is n-gram recall between the candidate and the set of references. \\
 \\
 $ROUGE_L = F_{LCS} = \frac{(1+\beta^2)R_{LCS}P_{LCS}}{R_{LCS}+\beta^{2}P_{LCS}}$ \\
 where $R_{LCS} = \frac{LCS(gen, ref)}{Len(ref)}$ and $P_{LCS} = \frac{LCS(gen, ref)}{Len(gen)}$ \\
 $LCS$ is the longest common subsequence and $Len$ is the length function.\\
 For multiple references: $ROUGE_{multi} = max_i(ROUGE(reference_i, candidate))$ \\ \hline

    \textbf{PINC}\cite{chen2011collecting} - Paraphrase In N-gram Changes \\
PINC is a measure of lexical dissimilarity and is calculated as the number of n-gram differences between the 
candidate and reference. \\
$PINC = \frac{1}{N}\sum_{n=1}^{N}{1 - \frac{|n-gram_{ref} \cap n-gram_{gen}|}{|n-gram_{gen}|}}$ \\
Candidates are rewarded for introducing new n-grams but not for omitting n-grams from the reference sentence. \\ \hline

    \textbf{PEM}\cite{liu2010pem} - Paraphrase Evaluation Metric\\ 
Metric based on adequacy, fluency and lexical dissimilarity. Adequacy is calculated independent of lexical (n-gram) similarity and 
fluency is calculated as $P_n  = \frac{logPr(S)}{length(S)}$ where Pr(S) is sentence probability predicted by a 
standard 4-gram language model. The three components are combined using SVM with radial basis function (RBF) kernel 
trained on human-judged paraphrase pairs. \\ \hline

    \textbf{TER}\cite{snover2006study} - Translation Edit Rate\\ 
Minimum number of edit required to change the candidate into one of the references, 
normalized by the average length of the reference. \\
$ TER = \frac{Num\ of\ edits}{Avg\ number\ of\ ref\ words}$ \\
All edits such as insertion, deletion, shifts and substitution have equal cost. \\\hline

    \textbf{Embedding based metrics} \cite{cer2018universal, sharma2017relevance}  \\ 
Calculate cosine similarity between candidate and reference sentence embedding. There are various ways to calculate sentence embeddings- \\
Word embedding average: $\bar{e_C} = \frac{\sum_{w \in C}{e_W}}{|\sum_{w \in C}{e_W}|}$ \\
Generate sentence embedding using a language model such as BERT \cite{devlin2018bert}.\\\hline

    \end{tabular}
\end{center}
\label{pop:metrics}
\end{table*}

\section{Model Fine-tuning and Generation}
\subsection{Fine-tuning GPT-2 for Paraphrasing}
\label{app:a}
GPT-2 was trained in an unsupervised manner on a dataset of 8 million webpages for the task of next word prediction. 
Naturally it is very adept at generational tasks. GPT-2 is a decoder only transformer and comes in small, medium, large and 
x-large sizes \cite{https://doi.org/10.48550/arxiv.1910.03771}, where the embedding dimension and the number of decoder blocks stacked over each other vary. Each decoder block consists of 
a masked self-attention layer and a feed-forward neural network layer. The mask in the attention layer is used to attend on tokens 
to the left, making the model uni-directional (left-to-right). Token here refers to words or sub-words the input sentence is 
split into using byte-pair encoding \cite{sennrich2015neural}. We encourage the reader to go through these articles for an introduction to 
transformers\footnote{https://jalammar.github.io/illustrated-transformer/} and GPT-2\footnote{https://jalammar.github.io/illustrated-gpt2/}

\begin{figure*}
  \caption{Input structure for fine-tuning. Padding is done on the right and is not attended upon 
  by the model. [PAD] is set as the same as [EOS] token. [SEP] token is added to the model vocabulary and is fine-tuned along with other token embeddings.}
  \centering
    \includegraphics[]{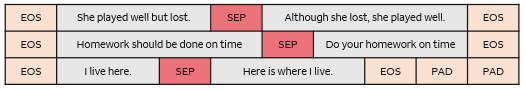}
  \label{gpt2:input}
\end{figure*}

Output text in GPT-2 is produced one token at a time, auto-regressively; that is, at each time step a token will be produced using previous tokens as context. 
Text generation from language models can either be open-ended or directed, 
where the latter implies an output which is a constrained transformation of the input \cite{holtzman2019curious}. 
Since we are dealing with directed generation, we also need to keep the input sentence in the context of each token generation step. 
As GPT-2 is a decoder only model, we provide the input sentence directly to the decoder for context. 
To fine-tune the model on a paraphrasing dataset, for each paraphrase pair (S1, S2), we input S1 into the 
model in expectation that it will generate S2. The input to the model for fine-tuning is structured as 
shown in \Cref{gpt2:input}, where S1 and S2 are separated by a special token- [SEP]. 
The whole structure is surrounded by [EOS] token, and excess length is filled by [PAD] token. 

\begin{figure*}
  \caption{Fine-tuning GPT-2 for paraphrasing based on maximum likelihood estimation. At a time-step, the model is expected to predict the next token of the sequence shown in  \Cref{gpt2:input}, starting from the first. Here, we show two such time-steps after the model reaches the [SEP] token. The loss accumulates from the [SEP] token till the [EOS] token is encountered in a sequence. The process continues for a pre-decided length (taken as 100 tokens here).}
  \centering
    \includegraphics[]{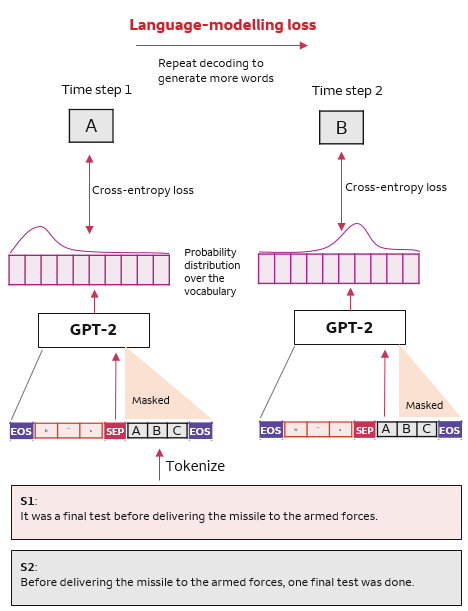}
  \label{gpt2:train}
\end{figure*}

GPT-2 starts from the first token of the sequence shown in \Cref{gpt2:input} and auto-regressively 
tries to predict the next one. We hide the future tokens using masking. At each time-step, 
GPT-2 generates a probability distribution $P$ over its internal vocabulary based on the token 
at the previous time step and all the preceeding ones. The loss at each time step is calculated as 
the cross-entropy between the probability distribution produced and the one-hot vector of the next token in the sequence. 
As we are only interested in making the model learn how to predict S2 given S1, 
we discount all loss arising from the predictions before the [SEP] token. \Cref{gpt2:train} shows the model training for two time-steps. 
The tokenized form of output sentence is written as \textit{A B C} for ease of illustration. At time-step 2, the model takes all tokens 
before including \textit{A} as the input and is expected to produce \textit{B} as the output. The loss at that time step is calculated as the 
cross-entropy between the probability distribution produced and the one-hot vector of token \textit{B} in the model vocabulary.

\begin{equation}
P(y_{1:n}) = \prod_{i=1}^{n}{P(y_{i}|y_{1:i-1}, x_{1:m})}
\label{sent:prob}
\end{equation}

Where $x_{1:m}$ are the input sentence tokens and $y_{1:n}$ are the output sentence tokens. 
The loss at time-step $t$ can be written as-

\begin{equation}
Loss_{t} = - \sum_{i=1}^{len(V)}{\hat{y}_{i}log(P_{t}(y_i))}
\end{equation}

Where $\hat{y}$ is the one-hot vector of the expected token and $len(V)$ is the length of the model vocabulary. 
This reduces to $Loss_{t} = -log(P_{t}(k)))$ where k is the expected token index at time t.

\subsection{Generating Paraphrases using GPT-2}
\label{app:b}

\begin{figure*}
  \caption{Structuring input for batch generation using GPT2. Notice that padding is done on the left so that 
  the [SEP] tokens align on the right for synchronous batch generation. This input pattern of [EOS]S1[SEP] 
  is recognized by the model from its fine-tuning (see \Cref{gpt2:input}).}
  \centering
    \includegraphics[]{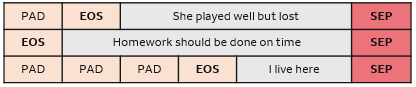}
  \label{gpt2:batch:inf}
\end{figure*}

\begin{figure*}
  \caption{Paraphrase generation using GPT-2. The diagram shown is for sequential decoding (and not beam search). 
  A token is sampled at each time-step from the generated probability distribution over the vocabulary of the model.
  The methodology is similar to \Cref{gpt2:train} except that the token generated at a time-step is appended to the input of the next.
  The process is halted if an [EOS] token is encountered or a limit of 50 tokens is reached.}
  \centering
    \includegraphics[]{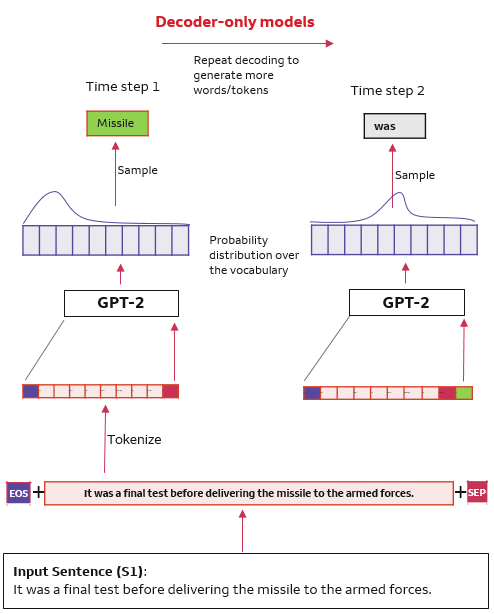}
  \label{gpt2:gen}
\end{figure*}

Generating paraphrases for an input sentence follows a similar pattern to that of fine-tuning the model. 
Since each input sentence is of different length, for batch-generation we pad the sentences on the left, as shown in \Cref{gpt2:batch:inf}.
The fine-tuning trains the model to generate a paraphrase for S1 when it sees the [SEP] token after it.
As before, at each step the model produces a probability 
distribution over its vocabulary and a token is sampled from it as the output. 
As shown in \Cref{gpt2:gen}, after the green token is generated in the first time step, 
it is appended to the input of the model at the next time step.
We keep generating tokens till a pre-decided length and truncate the output after [EOS] token is generated, which 
the model learns in its fine-tuning. 

We refer to decoding as the complete end-to-end process of generating sentences for a given model and an input.
Sampling refers to the way in which the output token is picked from the model generated probability distribution at a particular time-step.
Several decoding methods exist and a good article on decoding for open-ended 
text generation can be found in \citet{holtzman2019curious}. 
As each generated token from the model will have a probability associated with it, 
sentence probability can be calculated using \Cref{sent:prob}. It is not tractable to find the 
the optimum sequence with the highest probability for any generation task, including paraphrasing. 
Hence, beam search is often used as a method of maximization-based decoding, 
that is to search for sentences with the maximum probabilities. 
For open-ended generation, \citet{holtzman2019curious} 
discuss that beam search does not yield high quality text, and often results 
in repetition. Based on our observation and as also noted by the paper, this is usually not a problem for 
directed text generation as the output is tightly scoped to input.
Beam search works by storing a 
select number of beams (that is sentences) with the maximum probability at every time-step of the decoding process.
The number of beams to store is decided by the chosen beam-size. The model 
generates a probability distribution for each beam resulting from the previous decoding step. 
To find new beams for the current time-step, each probability distribution is 
shifted up by their respective beam probabilities, and new beams with the highest probabilities among them are selected. 
The last token in these new beams is a result of indirect sampling from the distributions generated in the last step.

\begin{figure*}
  \caption{Beam search as an approximate method to search for most likely sentences. Here, beam size is taken as 3. At each step, the model maintains 3 beams, that is 3 sequences of tokens accumulated that have had the highest probability in its search space. At the next step, the model generates 3 different probability distributions based on the 3 beams it gets. This results is vocabulary size times 3 number of sentence possibilities, from which the best 3 are picked again.}
  \centering
    \includegraphics[]{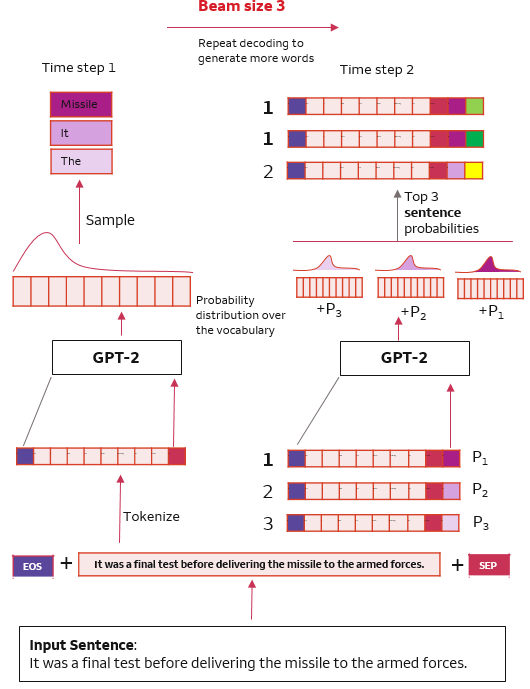}
  \label{beam:search}
\end{figure*}

\Cref{beam:search} shows the process graphically for two time-steps. Notice that 
at each step, we store 3 sentences with the highest probability. At time-step 1,
that corresponds to the 3 highest probability tokens, but at time-step 2 it is the cumulative 
probability that is considered. It could happen that the resultant beams spawn from a single beam, 
as shown in the figure. 

The alternative to maximization-based 
decoding is to disregard the sentence probability and sample individual tokens from the distribution at each time-step.
Several methods have been suggested in literature which improve our chances of sampling the right words by modifying the 
underlying distribution $P$, both for beam search and for sequential sampling of tokens. \Cref{sampling:tech} summarizes some of them.

\subsection{Paraphrase Samples}
\label{para:samp}
Here, we show a few paraphrase samples for generation settings referenced in the text. 
\Cref{mscoco:samples} shows some samples for the MSCOCO captions dataset generated using greedy and top-p decoding.
\Cref{msr:greedy:samples} and \Cref{msr:cand:samples} show sample generation for the MSR dataset using greedy sampling and top-p sampling followed by candidate selection respectively. Clear distinction can be made in the level of novelty between both the generation settings.
Finally, we also present reversed and shortened paraphrase samples for MSR in \Cref{msr:beam:inv:samples} and \Cref{msr:beam:short:samples} respectively.
More details are provided in the captions.

\FloatBarrier

\FloatBarrier
\begin{table*}
\caption{Description of popular sampling techniques. These methods modify the probability distribution generated by the model to 
increase the chance of a randomly picked token being desirable.}

\begin{center}
    \begin{tabular}{| p{15cm} |}
    \hline
    Method \\ \hline

    \textbf{Top-k} \\ 
	Before sampling, $k$ tokens with the highest probability are picked, and the resulting distribution is renormalized. 
	Now the output token is sampled from this new distribution. Greedy is essentially top-k with k=1. \\
	$p' = \sum_{y_i \in V(k)}{P(y_{i}|y_{1:i-1}, x_{1:m})}$ \\
	Where $P$ is the generated distribution for an input $x_{1:m}$ at step $i$, and $V(k)$ 
	is the set of k tokens which maximize $p'$. Then the new distribution $P'$ can be written as - \\
	$P'(y_{i}|y_{1:i-1}, x_{1:m}) = \frac{P(y_{i}|y_{1:i-1}, x_{1:m})}{p'}$ if $y_i \in V(k)$, else 0. \\ \hline

    \textbf{Top-p}  \cite{holtzman2019curious}  \\ 
	Top-p is similar to top-k, except that $k$ is not fixed and is decided based on the minimum number of 
	tokens needed to reach the desired cumulative probability cut-off. This essentially removes the fat-tail from the distribution 
	preventing the model from sampling tokens which have very low probability. \\
	$V(p)$ is the smallest set such that $\sum_{y_i \in V(p)}{P(y_{i}|y_{1:i-1}, x_{1:m})} \geq p$, \\
	then $p' = \sum_{y_i \in V(p)}{P(y_{i}|y_{1:i-1}, x_{1:m})}$. \\ \hline

    \textbf{Temperature scaling} \\ 
	It could happen that the distributions generated are sharp with very few tokens of high probability. 
	In that case the diversity of the text produced will be low. To tackle that problem, the distribution can be flattened 
	using temperature scaling with a factor of $T \in (0, 1]$. 
	$P'(y_{i}|y_{1:i-1}, x_{1:m}) = \frac{exp(P(y_{i}|y_{1:i-1}, x_{1:m})/T)}{\sum_{j}{exp(P(y_{j}|y_{1:j-1}, x_{1:m})/T)}}$\\ \hline

    \textbf{Repetition penalty} \\ 
	GPT-2 often has a tendency to repeat words or phrases that have previously ocurred \cite{holtzman2019curious}. 
	This manifests as the probability of previously ocurring tokens being unnaturally high in the distributions produced. 
	To tackle this we can apply a penalty on the probability of tokens which have ocurred previously or within a specific window. \\
	The modified distribution is $P'(y_{i}|y_{1:i-1}, x_{1:m}) = \frac{P(y_{i}|y_{1:i-1}, x_{1:m})}{p'}$ \\
	Where $p' = R$ if $y_i \in y_{i-w-1:i-1}$, else 1. Here $R$ is the penalty to be applied on window of size $w$. \\ \hline

    \end{tabular}
\end{center}
\label{sampling:tech}
\end{table*}

\begin{table*}
\caption{Paraphrase samples for MSCOCO captions generated using GPT2 medium with greedy and top-p sampling.
MSCOCO captions is a dataset of 5 sentence sets, describing the same image, considered as paraphrases for our purposes.
The amount of variation within the dataset is significantly higher than that of MSR, as can be seen from the examples.
}

\begin{center}    
    \begin{tabular}{ | M{0.8in}| M{0.8in}|M{0.8in}|M{0.8in}|M{0.8in}|M{0.8in}|M{0.8in}|}
    \hline
	Input & Ref1&  Ref2&  Ref3&  Ref4& Greedy & Top-p\\ \hline
	a biker with a helmet on is standing behind two motorcycles&
	a man in a helmet is standing beside two parked motorcyced&
	a guy with a helmet on stands next to a motorcycle&
	a man is standing next to two motorcycles&
	a picture of a person in leather and helmet standing behind two motorcycles&
	a man standing next to a motorcycle in a parking lot&
	a man standing next to two motorcycles \\ \hline

	a crowd of people walking down a street next to a traffic light&
	many people stand at a street corner together&
	a busy city intersection filled with people walking&
	a large crowd of people standing an a street&
	a crowded street corner blocked off by street cones&
	a crowd of people walking down a street&
	a view of people on the street during a traffic competition \\ \hline

	a 787 jet airplane sitting on a runway at an airport&
	a fence sits around a plane on a runway&
	a white and blue 787 plane at the airport&
	a commercial jetliner on a runway at an airport&
	a blue and white large body plane on the runway&
	a large jetliner sitting on top of an airport tarmac&
	a white jetliner taking off from an airport \\ \hline
    \end{tabular}
\end{center}
\label{mscoco:samples}
\end{table*}

\begin{table*}
\caption{Paraphrase samples for MSR generated using GPT-2 medium with greedy sampling. As is apparent, large portions of the input sentences are common in the generated sentences as well, something that we refer to as partial parroting in the text. In such generation settings, many sentences can be found that are exact replicas of the input sentence as well.}

\begin{center}    
    \begin{tabular}{ | M{2in}| M{2in}| M{2in}|}
    \hline
	Input paraphrase S1 & Reference paraphrase S2& Generated paraphrase \\ \hline
	But skeptics are concerned about the ease with which vendors can use these hardware-based security features to set digital rights management policies. &
	But skeptics are concerned about the ease at which these hardware-based security features could be used to set digital rights management policies by vendors. & 
	But skeptics are concerned about the ease with which vendors can use these hardware-based security features to set digital rights management policies. \\ \hline

	The company added, "until more facts are presented, Lindows.com will not take a position as to the validity of the claims presented by either side." &
	"Until more facts are presented, Lindows.com will not take a position as to the validity of the claims presented by either side," Lindows said in a statement. &
	"Lindows.com will not take a position as to the validity of the claims presented by either side," the company said.\\ \hline

	Dotson was arrested July 21 after calling 911, saying he needed help because he was hearing voices, authorities said.&
	Authorities picked up Dotson on July 21 after he called 911, saying he needed help because he was hearing voices, authorities said.&
	Dotson, 31, was arrested July 21 after calling 911 and saying he needed help because he was hearing voices, authorities said.\\ \hline
    
    \end{tabular}
\end{center}
\label{msr:greedy:samples}
\end{table*}

\begin{table*}
\caption{Samples for MSR generated using GPT-2 medium with candidate selection and top-p. Sampling with top-p helps add more relevant diversity to the generation. Candidate selection helps us filter for sentences having the desired level of novelty from the input sentence.}
\vspace*{-4mm}
\begin{center}    
    \begin{tabular}{ | M{2in}| M{2in}| M{2in}|}
    \hline
	Input paraphrase S1& Reference paraphrase S2& Generated paraphrase \\ \hline
	But skeptics are concerned about the ease with which vendors can use these hardware-based security features to set digital rights management policies. &
	But skeptics are concerned about the ease at which these hardware-based security features could be used to set digital rights management policies by vendors. & 
	Some security experts question the ease with which vendors can use these security features to set digital rights management policies.\\ \hline

	The company added, "until more facts are presented, Lindows.com will not take a position as to the validity of the claims presented by either side." &
	"Until more facts are presented, Lindows.com will not take a position as to the validity of the claims presented by either side," Lindows said in a statement. &
	Lindows.com said "until more facts are presented," it won't take a position on the validity of the claims presented by either side.\\ \hline

	Dotson was arrested July 21 after calling 911, saying he needed help because he was hearing voices, authorities said.&
	Authorities picked up Dotson on July 21 after he called 911, saying he needed help because he was hearing voices, authorities said.&
	Dotson, 30, told investigators he needed help because he was hearing voices, authorities said.\\ \hline
    
    \end{tabular}
\end{center}
\label{msr:cand:samples}
\end{table*}

\begin{table*}
\caption{Reversed paraphrases for MSR generated using GPT-2 medium with beam search and greedy sampling. Examples shown here are cherry-picked and it is not necessary for all sentences to be reversed. }
\vspace*{-4mm}
\begin{center}    
    \begin{tabular}{ |m{2in}| m{2in}| m{2in}|}
    \hline
	Input paraphrase S1& Reference paraphrase S2& Generated paraphrase \\ \hline
	According to Tuesday's report, consumers' assessment of current conditions was less favourable than a month earlier.&
	Consumers' assessment of current conditions was less favorable than last month.&
	Consumers' assessment of current conditions was less favourable than a month earlier, the report said. \\ \hline

	In September, Hewlett-Packard signed a development and marketing deal with the company. &
	Four months later it signed a joint marketing agreement with Hewlett-Packard Co. &
	In late September, the company signed a development and marketing deal with Hewlett-Packard. \\ \hline

	Clijsters was simply too complete and powerful for the Spanish veteran Conchita Martínez in her quarterfinal, winning, 6-2, 6-1.&
	Clijsters was simply too powerful for Spanish veteran Conchita Martinez, winning 6-2, 6-1. &
	The 6-2, 6-1 performance by Clijsters in her quarterfinal against the Spanish veteran Conchita Martínez was simply too complete and powerful to overcome. \\ \hline
    
    \end{tabular}
\end{center}
\label{msr:beam:inv:samples}
\end{table*}

\begin{table*}
\caption{Shortened paraphrase samples for MSR generated using GPT-2 medium with beam search and greedy sampling. Here, beams are penalized if their length exceeds a certain percentage of the input sentence.}
\vspace*{-4mm}
\begin{center}    
    \begin{tabular}{ | m{2in}| m{2in}| m{2in}|}
    \hline
	Input paraphrase S1 & Reference paraphrase S2 & Generated paraphrase \\ \hline
	According to Tuesday's report, consumers' assessment of current conditions was less favourable than a month earlier.&
	Consumers' assessment of current conditions was less favorable than last month.&
	Consumers' assessment of current conditions was also less favourable than a month earlier. \\ \hline

	In September, Hewlett-Packard signed a development and marketing deal with the company. &
	Four months later it signed a joint marketing agreement with Hewlett-Packard Co. &
	H-P entered into a development and marketing agreement with the company. \\ \hline

    \end{tabular}
\end{center}
\label{msr:beam:short:samples}
\end{table*}

\end{document}